\documentclass[runningheads]{llncs}
\usepackage{graphicx}
\usepackage{amsmath,amssymb} 
\usepackage{floatrow}
\newfloatcommand{capbtabbox}{table}[][\FBwidth]

\usepackage{multirow}

\usepackage{color}
\usepackage{algorithm2e}
\usepackage[dvipsnames]{xcolor}
\usepackage{color}
\usepackage{enumitem}
\setlist{nolistsep}
\usepackage{makecell}

\usepackage{subcaption}
\usepackage{adjustbox}
\usepackage{amsmath}
\usepackage{amssymb}
\usepackage{array}
\usepackage{bm}
\usepackage{booktabs}
\usepackage{color}
\usepackage{epsfig}
\usepackage{graphicx}
\usepackage{multirow}
\usepackage{url}
\usepackage{adjustbox}
\usepackage{xspace}
\usepackage[pagebackref=true,breaklinks=true,colorlinks,bookmarks=false]{hyperref}

\usepackage{array}
\newcolumntype{L}[1]{>{\raggedright\let\newline\\\arraybackslash\hspace{0pt}}m{#1}}
\newcolumntype{C}[1]{>{\centering\let\newline\\\arraybackslash\hspace{0pt}}m{#1}}
\newcolumntype{R}[1]{>{\raggedleft\let\newline\\\arraybackslash\hspace{0pt}}m{#1}}

\makeatletter
\DeclareRobustCommand\onedot{\futurelet\@let@token\@onedot}
\def\@onedot{\ifx\@let@token.\else.\null\fi\xspace}

\usepackage{lipsum}

\newcommand\blfootnote[1]{%
  \begingroup
  \renewcommand\thefootnote{}\footnote{#1}%
  \addtocounter{footnote}{-1}%
  \endgroup
}

\newcommand{\etal}{\textit{et al}.}
\usepackage[width=122mm,left=12mm,paperwidth=146mm,height=193mm,top=12mm,paperheight=217mm]{geometry}
\begin{document}

\pagestyle{headings}
\mainmatter
\def\ECCV18SubNumber{11}  

\title{Object Pose Estimation from Monocular Image using Multi-View Keypoint Correspondence} 



\author{Jogendra Nath Kundu*  \and
Rahul M V*   \and
Aditya Ganeshan* \and
R Venkatesh Babu}
\institute{Indian Institute of Science, Bengaluru, India}
%
%

\maketitle

\begin{abstract}
\blfootnote{* denotes equal contribution.}
Understanding the geometry and pose of objects in 2D images is a fundamental necessity for a wide range of real world applications. Driven by deep neural networks, recent methods have brought significant improvements to object pose estimation. However, they suffer due to scarcity of keypoint/pose-annotated real images and hence can not exploit the object's 3D structural information effectively. In this work, we propose a data-efficient method which utilizes the geometric regularity of intraclass objects for pose estimation. First, we learn pose-invariant local descriptors of object parts from simple 2D RGB images. These descriptors, along with keypoints obtained from renders of a fixed 3D template model are then used to generate keypoint correspondence maps for a given monocular real image. Finally, a pose estimation network predicts 3D pose of the object using these correspondence maps. This pipeline is further extended to a multi-view approach, which assimilates keypoint information from correspondence sets generated from multiple views of the 3D template model. Fusion of multi-view information significantly improves geometric comprehension of the system which in turn enhances the pose estimation performance. Furthermore, use of correspondence framework responsible for the learning of pose invariant keypoint descriptor also allows us to effectively alleviate the data-scarcity problem. This enables our method to achieve \textit{state-of-the-art} performance on multiple real-image viewpoint estimation datasets, such as Pascal3D+ and ObjectNet3D. To encourage reproducible research, we have released the codes for our proposed approach. \footnote{\href{https://github.com/val-iisc/pose\_estimation}{https\://github.com/val\-iisc/pose\_estimation}. } 
\end{abstract}

\section{Introduction}

Estimating 3D pose of an object from a given RGB image is an important and challenging task in computer vision. Pose estimation can enable AI systems to gain 3D understanding of the world from simple monocular projections. 
While ample variation is observed in the design of objects of a certain type, say chairs, the intrinsic structure or skeleton is observed to be mostly similar. Moreover, in case of 3D objects, it is often possible to unite information from multiple 2D views, which in succession can enhance 3D perception of humans as well as artificial vision systems. In this work, we show how intraclass structural similarity of objects along with multi-view 3D interpretation can be utilized to solve the task of fine-grained 3D pose estimation.

By viewing instances of an object class from multiple viewpoints over time, humans gain the ability to recognize sub-parts of the object, independent of pose and intra-class variations. Such viewpoint and appearance invariant comprehension enables human brain to match semantic sub-parts between different instances of same object category, even from simple 2D perspective projections (RGB image). Inspired from human cognition, an artificial model with similar matching mechanism can be designed to improve final pose estimation results. In this work, we consider a single template model with known keypoint annotations as a 3D structural reference for the object category of interest. Subsequently, Key-point correspondence maps are obtained by matching keypoint-descriptors of synthetic RGB projections from multiple viewpoints, with respect to the spatial descriptors from a real RGB image. Such keypoint-correspondence maps can provide the geometric and structural cues useful for pose estimation. 

\begin{figure*}[!t]
\centering    
	\includegraphics[width=0.95\linewidth]{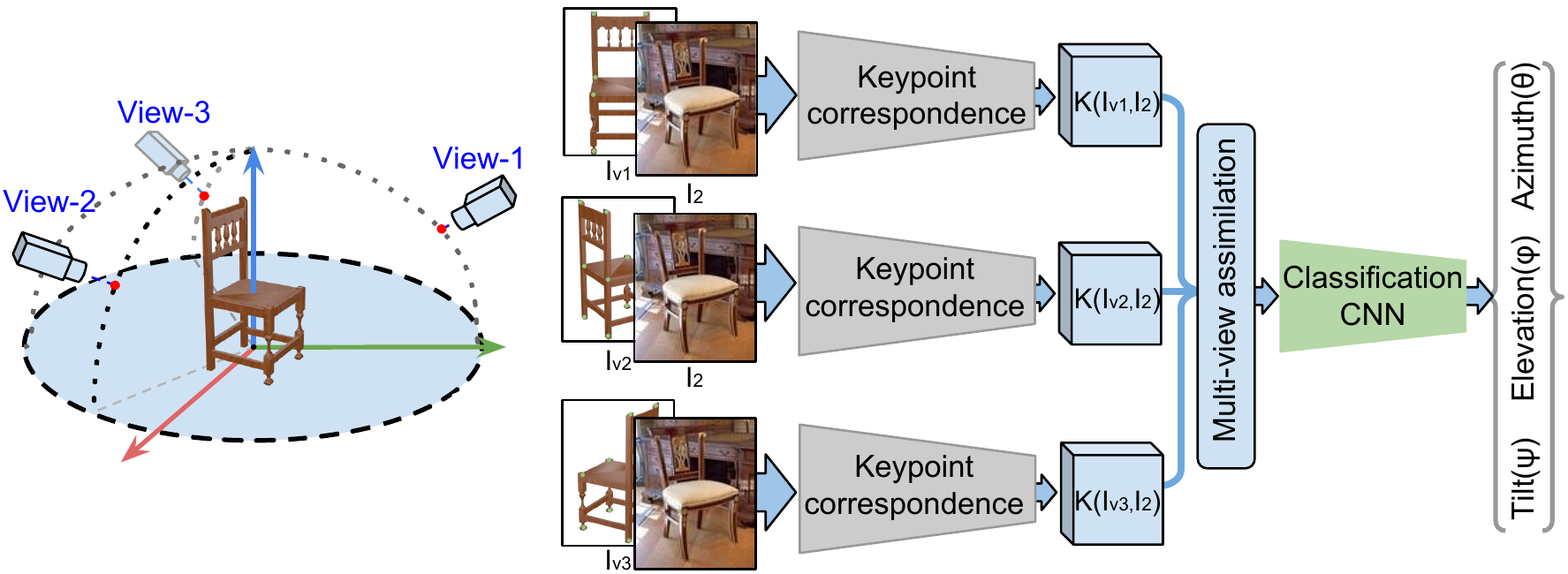}
	\caption{Illustration of the proposed pipeline. 
	Given a real image $I_2$, it is paired with multiple 2D views of a template 3D model with annotated keypoints. For each pair of images, keypoint correspondence maps are generated, represented by $K(I_{vk},I_2)$. Finally, the pose estimator network assimilates information from all correspondence maps to predicts the pose parameters.  
	} 
	\label{fig:fig_1}    
\end{figure*}

The proposed pose estimation system consists of two major parts; 1) A Fully Convolutional Network which learns pose-invariant local descriptors to obtain keypoint-correspondence, and 2) A pose estimation network which fuses information from multiple correspondence maps to output the final pose estimation result. For each object class, we annotate a single template 3D model with sparse 3D keypoints. Given an image, in which the object's pose is to be estimated, first it is paired with multiple rendered images from different viewpoints of the template 3D model  (see Figure \ref{fig:fig_1}). Projections of the annotated 3D keypoints is tracked on the rendered synthetic images to provide ground-truth for learning of efficient key-point descriptor. Subsequently, keypoint-correspondence maps are generated for each image pair using correlation of individual keypoint descriptor (from rendered image) to the spatial descriptors obtained from the given image. 

Recent works \cite{mvRotationNet,mvtriplet,mvhumanpose} show that deep neural networks can effectively merge information from multiple 2D views to deliver enhanced view estimation performance. These approaches require multi-view projections of the given input image to exploit the muli-view information. But in the proposed approach, we attempt to take advantages of multi-view cue by generating correspondence map from the single-view real RGB image by comparing it against multiview synthetic renders. This is achieved by feeding the multi-view keypoint correspondence maps through a carefully designed fusion network (convolutional neural network) to obtain the final pose estimation results. Moreover, by fusing information from multiple viewpoints, we show significant improvement in pose estimation, making our pose estimation approach \textit{state-of-the-art} in competitive real-image datasets, such as Pascal3D+ \cite{xiang2014beyond} and ObjectNet3D \cite{xiang2016objectnet3d}.  In Figure \ref{fig:fig_1}, a diagrammatic overview of our approach is presented.


Many recent works \cite{tulsiani2015viewpoints,xiang2014beyond,Grabner18}, have utilized deep neural networks for 3D object understanding and pose estimation. However, these approaches have several drawbacks. Works such as \cite{su2015render,wu2016single} achieve improved pose estimation performance by utilizing a vast amount of synthetic data. This can be a severe bottleneck when an extensive repository of diverse 3D models for a specific category is unavailable (as in case of novel object-classes, such as mechanical parts, abstract 3D models etc.). Additionally, 3D-INN~\cite{wu2016single} require a complex keypoint-refinement module that, while being remarkable at keypoint estimation, shows sub-optimal performance for viewpoint estimation, when compared against current state-of-the-art models. We posit that it is essential to explore and exploit strong 3D-structural object priors to alleviate various general issues, such as data-bottleneck and partial-occlusion, which are observed in object viewpoint estimation. Moreover, our approach has two crucial advantages. Firstly, our keypoint correspondence map captures relation between the keypoint and the entire 2D spatial view of the object in a given image. That is,  the correspondence map not only captures information regarding spatial location of keypoint in the given image, but also captures various relations between the keypoint and other sematic-parts of the object. In Figure \ref{fig:fig_2}, we show the obtained correspondence map for varied keypoints, and provide evidence for this line of reasoning. Secondly, our network fuses the correspondence map of each keypoint from multiple views. This multi-view comprehension of individual keypoint enables our network to have a more nuanced interpretation of 3D structure of the object class, which later leads to improvement in pose estimation performance. 
 

To summarize, our main contributions in this work include: (1) a method for learning pose-invariant local descriptors for various object classes, (2) a keypoint correspondence map formulation which captures various explicit and implicit relations between the keypoint, and a given image, (3) a pose estimation network which assimilates information from multiple viewpoints, and (4) state-of-the-art performance on real-image object pose estimation datasets for indoor object classes such as `Chair', `Sofa', `Table' and `Bed'.

\section{Related work}

\textbf{Local descriptors and keypoint correspondence:} A multitude of work propose formulations for local discriptors of 3D objects, as well as 2D images. Early methods employed hand-engineered local descriptors like SIFT or HOG \cite{aubry2014seeing,liu2016sift,taniai2016joint,berg2005shape} to represent semantic part structures useful for object comprehension. With the advent of deep learning, works such as \cite{schmidt2017self,han2017scnet,yu2018hierarchical,choy2016universal} have proposed effective learning methods to obtain local descriptor correspondence in 2D images. Recently, Huang \etal~\cite{Huang:2017:LMVCNN} propose to learn local descriptors for 3D objects following deep multi-view fusion approach.  While this work is one of our inspirations, our method differs in many crucial aspects. We do not require extensive multi-view fusion of local descriptors as performed by Huang \etal   for individual local points. Moreover we do not rely on a large repository of 3D models with surface segmentation information for generalization. For effective local descriptor correspondence, Universal Correspondence Network \cite{choy2016universal} formulate an optimization strategy for learning robust spatial correspondence, which is used in coherence with an active hard-mining strategy and a convolutional spatial transformer (STN) . While \cite{choy2016universal} learn geometric and spatial correspondence for task such as semantic part matching, we focus on the learning procedure of their approach and adapt it for learning our pose-invariant local descriptors.\\

\noindent
\textbf{Multi-view information assimilation :} Borotschnig \etal~\cite{mv_1}, and Paletta \etal~\cite{mv_2} were one of the earliest works to show the utility of multi-view information for improving performance on tasks related to 3D object comprehension. In recent years, multiple innovative network architectures, such as \cite{mvtriplet,mvhumanpose} have been proposed for the same. One of the earliest works to combine deep learning with multi-view information assimilation, \cite{MVCNN} showed that 2D image-based approaches are effective for general object recognition tasks, even for 3D models. They proposed an approach for 3D object recognition based on multiple 2D projections of the object, surpassing previous works which were based on other 3D object representations such as voxel and mesh format. In \cite{mvOverview}, Qi \etal give a comprehensive study on the voxel-based CNN and multi-view CNN for 3D object classification. 
Apart from object classification, multi-view approach is seen to be useful for a wide variety of other tasks, such as learning local features for 3D models \cite{Huang:2017:LMVCNN}, 3D object shape prediction \cite{mvcTulsiani18} etc.. In this work, we use multi-view information assimilation for object pose estimation in a given monocular RGB image using multiple views of a 3D template model. Such a multi-view approach does not exist in the literature.\\

\noindent
\textbf{Object viewpoint estimation:} Many recent works \cite{poirson2016fast,mahendran20173d}  use deep convolutional networks for object viewpoint estimation. While works such as \cite{tulsiani2015viewpoints} attempt pose estimation along with keypoint estimation, an end-to-end approach solely for 3D pose estimation was first proposed by RenderForCNN \cite{su2015render}. Su \etal~\cite{su2015render} proposed to utilize vast amount of synthetic rendered data from 3D CAD models with dataset specific cues for occlusion and clutter information, to combat the lack of pose annotated real data. In contrast, 3D Interpreter Network (3D-INN) \cite{wu2016single} propose an interesting approach where 3D keypoints and view is approximated by minimizing a novel re-projection loss on the estimated 2D keypoints. However, the requirement of vast amount of synthetic data is a significant bottleneck for both the works. In comparison, our method 
relies on the presence of a single synthetic template model per object category, making our method significantly data efficient and far more scalable. This is an important pre-requisite to incorporate the proposed approach for novel object classes, where multiple 3D models may not exists. Recently, Grabner~\etal~\cite{Grabner18} estimate object pose by predicting the vertices of a 3D bounding box and solving a perspective-n-point problem. While achieving state-of-the-art performance in multiple object categories, they could not surpass performance of \cite{su2015render} on the challenging indoor object classes such as `chair',`sofa', and `table'. It is essential to provide stronger 3D structural priors to learn pose estimation under data scarcity scenario for such complex categories. The structural prior is effectively modeled in our case by keypoint correspondence and multi-view information assimilation.

\section{Approach}

This section consist of 3 main parts: in Section \ref{sub:1}, we present our approach for learning pose invariant local descriptors, Section \ref{sub:2} explains how the keypoint correspondence maps are generated, and Section \ref{sub:3} explains our regression network, along with various related design choices. Finally, we briefly describe our data generation pipeline in Section \ref{sub:4}.

\subsection{Pose-Invariant Local Descriptors }
\label{sub:1}

To effectively compare given image descriptors with the keypoint descriptors from multi-view synthetic images, our method must identify various sub-parts of the given object, invariant to pose and intra-class variation. To achieve this we train a convolutional neural network (CNN), which takes an RGB image as input and gives a spatial map of local descriptors as output.  That is, given an image $I_1$ of size $h \times w$, our network predicts a spatial local descriptor map $L_{I_1}$ of size $h \times w \times d$ , where the $d$-dimensional vector at each spatial location is treated as the corresponding local descriptor.

 Following the approach of other established method \cite{Huang:2017:LMVCNN,choy2016universal}, we use the CNN to form two brances of a Siamese architecture with shared convolutional parameters. Now, given a pair of images $I_1$ and $I_2$ with annotated keypoints,  we pass them through the siamese network to get the spatial local descriptor maps $L_{I_1} $ and $L_{I_2}$ respectively. The annotated keypoints are then used to generate positive and negative correspondence pairs, where a positive correspondence pair refers to a pair of points $I_1(x_k, y_k), I_2({x'}_{k}, {y'}_{k})$ such that they represent a certain semantic keypoint. 
In \cite{choy2016universal}, authors present the correspondence contrastive loss, which is used to reduce the distance between the local descriptors of positive correspondence pairs, and increase the distance for the negative pairs. Let  $\mathbf{x_i}= (x_k, y_k)$ and $ \mathbf{x'_i} = ({x'}_{k}, {y'}_{k}) $ represent spatial locations on $I_1$ and $I_2$ respectively. The correspondence contrastive loss can be defined as,

\begin{align}\label{eqn:1}
\begin{split}
Loss =
	\dfrac{1}{2N}\sum_{i}^{N} & \big\{ s_i{\lVert L_{I_1}(\mathbf{x}) - L_{I_2}(\mathbf{x'}) \rVert}^2 + \\
	&( 1-s_i) \max{(0, \; m-{\lVert L_{I_1}(\mathbf{x}) - L_{I_2}(\mathbf{x'}) \rVert}^2)  }\big\}
\end{split}
\end{align}

\noindent
where $N$ is the total number of pairs, $s_i = 1$ for positive correspondence pairs, and $s_i = 0$ for negative correspondence pairs.

Chief benefit of using a correspondence network is its utility to combat data-scarcity. Given $N$ samples with keypoint annotation, we can generate $^NC_2$ training samples for training the local descriptor representations. The learned local descriptors do most of the heavy lifting by providing useful structural cues for 3D pose estimation. This helps us avoid extensive usage of synthetic data and the common pitfalls associated with it, such as domain shift~\cite{kundu2018adadepth} while testing on real samples.  Compared to state-of-the-art works ~\cite{su2015render,wu2016single}, where millions of synthetic data samples were used for effecting training, we use only $8k$ renders of a single template 3D model per class (which is less than $1\%$ of the data used by \cite{su2015render,wu2016single}). Another computational advantage we observe is in terms of run-time efficiency. Given a single image, we estimate the local descriptors for all the visible points on the object. This is in stark contrast to Huang \etal~\cite{Huang:2017:LMVCNN}, where multiple images were used for generating local descriptors for each point of the object.

In most cases such as in \cite{wu2016single}, objects are represented by a sparse set of keypoints. Learning feature descriptors for only a few sparse semantic keypoints has many disadvantages. In such case, the models fails to learn efficient descriptors for spatial regions away from the defined semantic keypoint locations. However, information regarding parts away from these keypoints can also be useful for pose estimation. Hence, we propose to learn proxy-dense local descriptors to obtain more effective correspondence maps (see Figure \ref{fig:data}b and \ref{fig:data}c). This also allows us to train the network more efficiently by generating enough amount of positive and negatives correspondence pairs. For achieving this objective, we generate dense keypoints for all images, details of which are presented in Section~\ref{sub:3}.

\begin{figure*}[!t]
\centering    
	\includegraphics[width=0.95\linewidth]{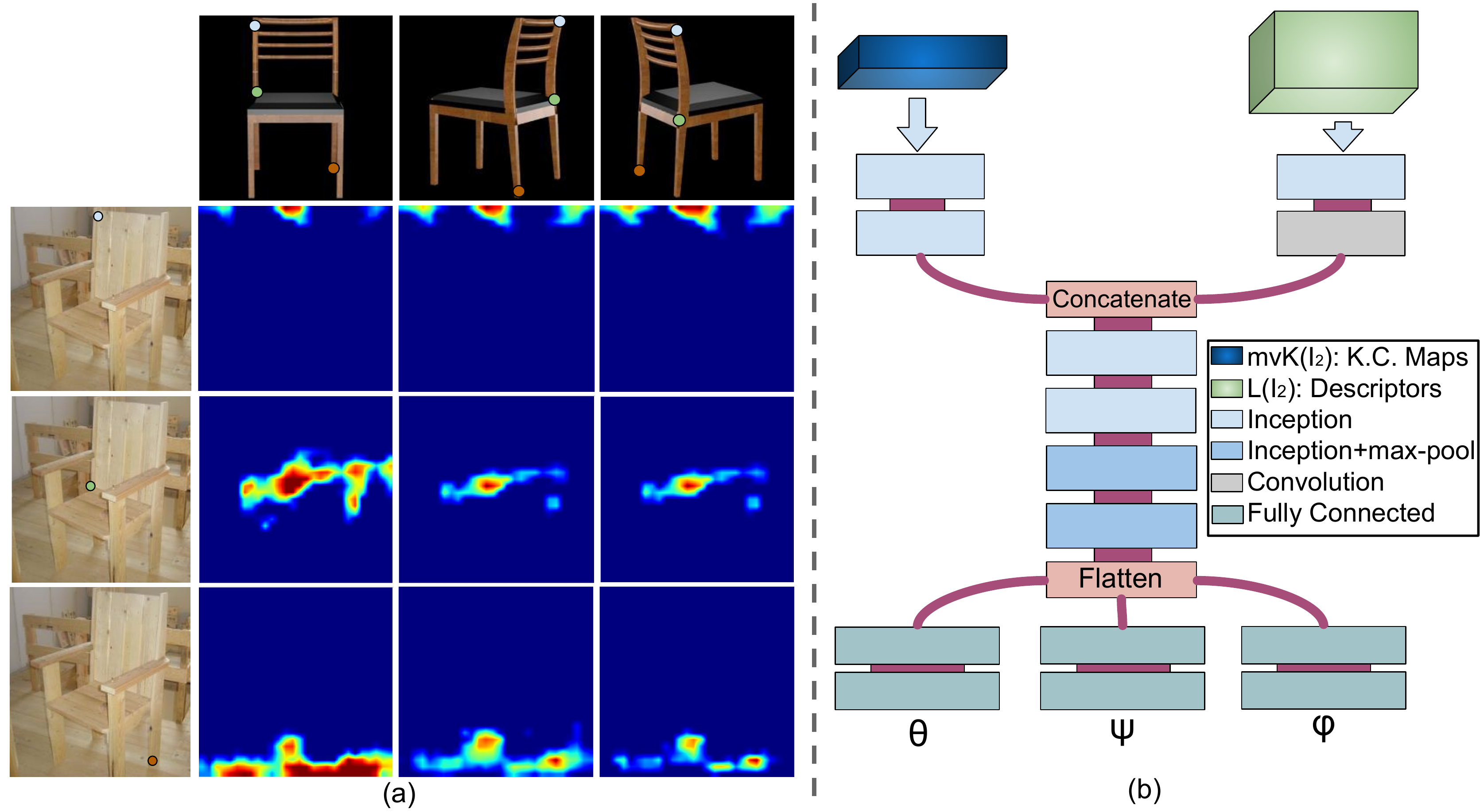}
	\caption{(a) The Keypoint Correspondence map generated by our approach. The top row shows the template 3D model from $3$ Views where $3$ different keypoints are highlighted. First column shows the real image where pose has to be estimated. As we can see, Keypoints have lesser ambiguity when looked from views where they are clearly visible (For eg., back-leg keypoint, View 2 and 3). (b) The architecture of our pose estimator network.} 
	\label{fig:fig_2}    
\end{figure*}

\noindent
\textbf{Correspondence Network Architecture:} \hspace{1mm}
The siamese network contains two branches with shared weights. It is trained on the generated key-point annotations (details in section~\ref{sub:3}) using the loss, equation~\ref{eqn:1} described above. For the Siamese network, we employ a standard Googlenet~\cite{szegedy2015going} architecture with imagenet pretrained weights. Further, to obtain spatially aligned local features $L_I$, we use a convolutional spatial transformation layer after $pool4$ layer of googlenet architecture, as proposed in UCN \cite{choy2016universal}. The use of convolutional spatial transformation layer is found to be very useful for semantic part correspondence in presence of reasonably high pose and intra-class variations. 

\subsection{Keypoint Correspondence Maps}
\label{sub:2}

The CNN introduced in the previous section provides a spatial local descriptor map $L_{I_1}$ for a rendered synthetic image $I_1$. Now, using the keypoint annotations rendered from the 3D template model, we want to generate a spatial map, which can capture the location of corresponding keypoint in a  given real image, $I_2$. To achieve this we propose to utilize pairwise descriptor correlation between both the images. Let, $L_{I_1}$ is of size $h \times w \times d$, and $x_k$ represents a keypoint in $I_1$. Now our goal is to estimate a correspondence map of keypoint $x_k$ for the real image $I_2$. 
By taking correlation of the local descriptor at $x_k$, $L_{I_1}(x_k)$ with all locations $(i',j')$ of the spatial local descriptor for image  $I_2$, i.e. $L_{I_2}$, correspondence maps are obtained for each keypoint, $x_k$. Using max-out Hadamard product $H$, we compute the pairwise descriptor correlation for any $(i',j')$ in $I_2$ and $x_k$ in $I_1$as follows:
$$
H(x_k, (i,j))  =  \max(0,L_{I_1}(x_k)^T L_{I_2}(i',j'))
$$
$$
C_{x_k,I_2}(L_{I_1}(x_k), \; L_{I_2}(i',j')) = \frac{\exp^{H(x_k,i,j)}}{\sum_{p, q} \exp^{H(x_k,p,q)}}
$$

As the learned local descriptors are unit normalized, the max-out Hadamard product $H(x_k, (i,j))$ represents only positive correlation between local descriptor at $x_k$ with local descriptors of all locations $(i,j)$ in image $I_2$. By applying softmax on the entire map of rectified Hadamard product, multiple high correlation values will be suppressed by making the highest correlation value more prominent in the final correspondence map. Such normalization step is in line with the traditionally used second nearest neighbor test proposed by Lowe \etal ~\cite{lowe2004distinctive}. 
Using the above formulation, keypoint correspondence maps $C_{x_k,I_2}$ is generated for a set of sparse structurally important keypoints $x_k, for k=1,2,...,N$ in image $I_1$. The structurally important keypoints that we use for each object class are the same as the ones used by \cite{wu2016single}. Finally, We use the structurally important keypoint set for individual object category as defined by Wu \etal~\cite{wu2016single}. Finally the stacked correspondence map for all structural keypoints of $I_1$ computed for image $I_2$ is represented by $K(I_1, I_2)$. Here $K(I_1, I_2)$ is of size $N \times h \times w$, where $N$ is the number of keypoints. 

As explained earlier, our keypoint correspondence map computes the relation between the keypoint $x_k$ in $I_1$ and all the points $(i,j)$ in $I_2$. In comparison to \cite{wu2016single}, where a location heatmap is predicted for each keypoint, our keypoint correspondence map captures the interplay between different keypoints as well. This in turn acts as an important cue for final pose estimation. Figure \ref{fig:fig_1} shows keypoint correspondence maps generated by our approach, which clearly provide evidence of our claims.

\subsection{Multi-view Pose Estimation Network}
\label{sub:3}
With the structural cues for object in image $I_2$ provided by the keypoint correspondence set $K(I_1,I_2)$, we can estimate pose of the object more effectively. In our setup, $I_1$ is a synthetically rendered image of the template 3D model with the tracked 2D keypoint annotations, and $I_2$ is the image of interest where the pose has to be estimated. It is important to note, that $K(I_1,I_2)$ contains information regarding relation between the keypoints $x_k,k=1,2,...,N$ in $I_1$ with respect to the image $I_2$. However, as $I_1$ is a 2D projection of the 3D template object, it is possible that some keypoints are self occluded, or only partially visible. For such keypoints $C_{x_k,I_2}$ would contain noisy and unclear correspondence. As mentioned earlier, the selected keypoints are structurally important and hence lack of information of any of them can hamper the final pose estimation performance.

To alleviate this issue, we propose to utilize a multi-view pose estimation approach. We first render the template 3D model from multiple viewpoints $I_{v1},I_{v2}, ... I_{vm}$ considering $m$ viewpoints. Then, the keypoint correspondence set is generated for each view by pairing $I_{vk}$ with $I_2$ for all $k$. Finally, information from multiple views is combined together by concatenating all the correspondence sets to form a fused Multi-View Correspondence set, represented by $mvK(I_2)$. Here, $mvK(I_2)$ is of size $(m \times N,h,w)$; where $m$ is the number of views, and $N$ is the number of structurally important keypoints.  subsequently, $mvK(I_2)$ is supplied as an input to our pose estimation network which effectively combines information from multiple-views of the template object to infer the required structural cues. For a given $m$ , we render $I_{v1},I_{v2}, ... I_{vm}$ from fixed viewpoints, $v_k = (360/m \times k, 10,0)$ for $k =1,2,...m$; where $v_k$ represents a tuple of azimuth, elevation and tilt angles in degree.

In Figure \ref{fig:fig_2}b, the architecture of our pose estimation network is outlined. Empirically, we found Inception Layer to be most efficient in terms of performance for memory footprint. We believe, multiple receptive fields in the inception layer help the network to learn structural relations at varied scales, which later improves pose estimation performance.  For effective modeling,  we consider deeper architecture with reduced number of filters per convolutional layer. Here, the pose estimation network classifies the three Euler angles, namely azimuth ($\theta$), elevation ($\phi$), and tilt ($\psi$). Following \cite{su2015render}, we use the Geometric Structure Aware Classification Loss for effective estimation of all the three angles.

As a result of proxy-dense correspondence, Pose-Invariant local descriptor $L(I_2)$ has information about dense keypoints. But $mvK(I_2)$ leverages information only from the sparse set of structurally important keypoints. Therefore, we also explore whether $L(I_2)$ can also be utilized to improve the final pose estimation performance. To achieve this, we concatenate convolution-processed feature map of $L(I_2)$ with inception-processed features of $mvK(I_2)$ to form the input to our pose-estimation network. This brings us to our final state-of-the-art architecture. Various experiments are performed in section \ref{experiments:1}, which outline the benefits of each of the design choices. 

\begin{figure*}[!t]
\centering    
	\includegraphics[width=0.95\linewidth]{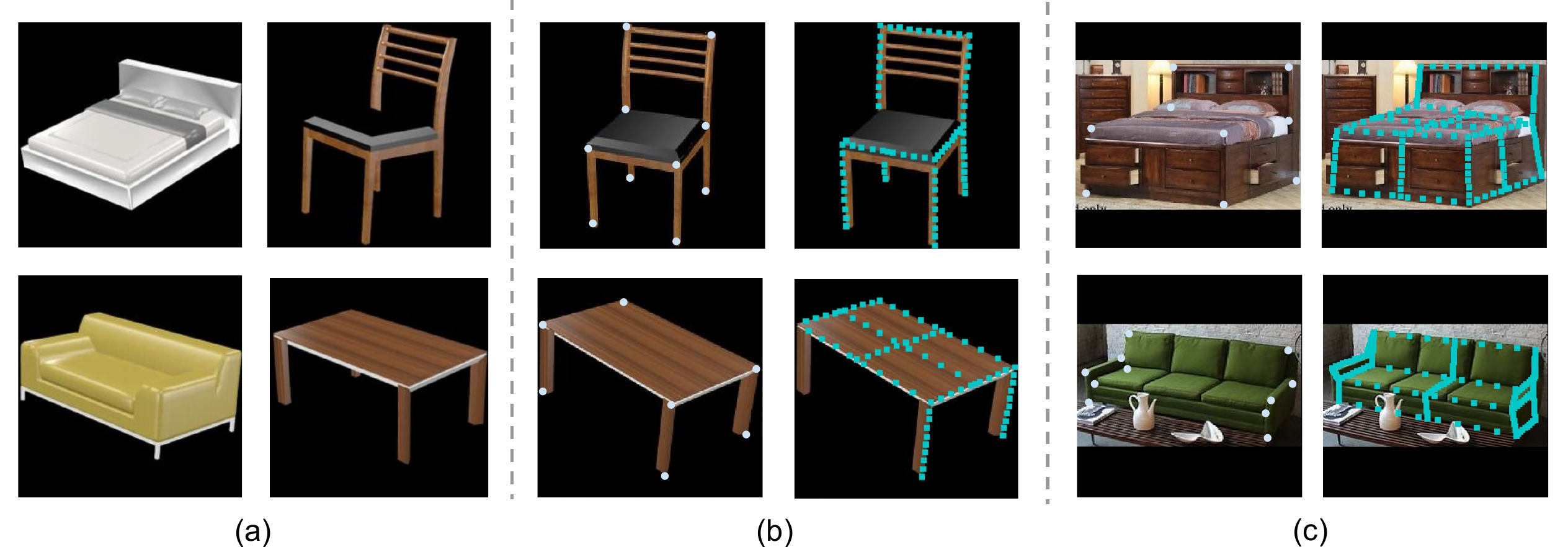}
	\caption{(a) The single 3D template model selected for each class. (b) Template models are annotated with sparse 3D keypoints, which are projected to 2D keypoints in each rendered image. From these keypoints, dense keypoint annotation is generated by sampling along the skeleton. (c) Similar process is used on real image datasets where sparse 2D keypoint annotation has been provided. 
	} 
	\label{fig:data}    
\end{figure*}

\subsection{Data Generation for Local Descriptors}
\label{sub:4}

Learning an efficient pose-invariant keypoint descriptor requires presence of ground-truth positive correspondence pair in sufficient amount. For each real image, we generate an ordered set of dense keypoints by forming a skeletal frame of the object from the available sparse keypoint annotations provided in Keypoint-5 dataset \cite{wu2016single}. To obtain dense positive keypoint pairs, we sample additional points along the structural skeleton lines obtained from the semantic sparse keypoints for both real and sythetic image. Various simple keypoint pruning methods based on seat presence, self-occlusion etc. are used to remove noisy keypoints (more detail in supplementary). Figure~\ref{fig:data} (c) shows some real images where dense keypoint annotation is generated from available sparse keypoint annotation as described above.

For our synthetic data, a single template 3D model (per category) is manually annotated with a sparse set of 3D keypoints. These models are shown in Figure~\ref{fig:data}a. Using a modified version of the rendering pipeline presented by \cite{su2015render}, we render the template 3D model and project sparse 2D keypoints from multiple views to generate synthetic data required for the pipeline. Similar skeletal point sampling mechanism as mentioned earlier is used to from dense keypoint annotation for each synthetic image as shown in Figure~\ref{fig:data}b (more details in supplementary). 

\section{Experiments}

In this section, we evaluate the proposed approach with other state-of-the-art models for multiple tasks related to viewpoint estimation. Additionally, multiple architectural choices are validated by performing various ablation on the proposed multi-view assimilation method. 

\noindent
\textbf{Datasets: } 
We empirically demonstrate \textit{state-of-the-art} or competitive performance when compared to several other methods on two public datasets. \textit{Pascal3D+}~\cite{xiang2014beyond}:  
This dataset contains images from Pascal~\cite{everingham2015pascal} and ImageNet~\cite{russakovsky2015imagenet} set labeled with both detection and continuous pose annotations for 12 rigid object categories. \textit{ObjectNet3D}~\cite{xiang2016objectnet3d}:
This dataset consists of 100 diverse categories, 90,127 images with 201,888 objects.
Due to the requirement of keypoints, keypoint-based methods can be evaluated only on object-categories with available keypoint annotation. Hence, we evaluate our method on 4 categories from these dataset namely, Chair, Bed, Sofa and Dining-table (3 on Pascal3D+, as it does not contain Bed category). 
We evaluate our performance for the task of object viewpoint estimation, and joint detection and viewpoint estimation. 

\noindent
 \textbf{Metrics}: Performance in object viewpoint estimation is measured using \textit{Median Error} ($MedErr$) and  \textit{Accuracy at $\theta$} ($Acc_\theta$), which were introduced by Tulsiani \etal~\cite{tulsiani2015viewpoints}. $MedErr$ measures the median geodesic distance between the predicted pose and the ground-truth pose (in degree) and  $Acc_{\theta}$ measures the $\%$ of images where the geodesic distance between the predicted pose and the ground-truth pose is less than $\theta$ (in radian). While previous works evaluate $Acc_\theta$ with $\theta=\pi/6$ only, we evaluate $Acc_\theta$ with smaller $\theta$ as well (i.e. for $\theta = \pi/8$ and $\pi/12$) to highlights our models ability to deliver more accurate pose estimates. %
Finally, to evaluate performance on joint detection and viewpoint estimation, we use \textit{Average Viewpoint Precision at `n' views}($AVP$-$n$) metric as introduced in \cite{xiang2014beyond}.

\noindent
\textbf{Training details: }
We use ADAM optimizer~\cite{kingma2014adam} having a learning rate of $0.001$ with minibatch-size $7$. For each object class, we assign a \textit{single} 3D model from Shapenet Repository as the object template. The local feature descriptor network is trained using 8,000 renders of the template 3D model (per class), along with real training images from Keypoint-5 and Pascal3D+. Dense correspondence annotations are generated for this segment of the training (refer Section~\ref{sub:4}). Finally, the pose estimation network is trained using Pascal3D+ or ObjectNet3D datasets. This training regime provides us our normal model, labeled $\mathbf{Ours_N}$. Additionally, to compare against RenderForCNN~\cite{su2015render} in the presence of synthetic data, we construct a separate training regime, where the synthetic data provided by RenderForCNN~\cite{su2015render} is also utilized for training the pose estimation network. The model trained in this regime is labeled $\mathbf{Ours_D}$.

  


\begin{figure}[t]
\begin{floatrow}
\ffigbox{%
  
	\includegraphics[width=\linewidth]{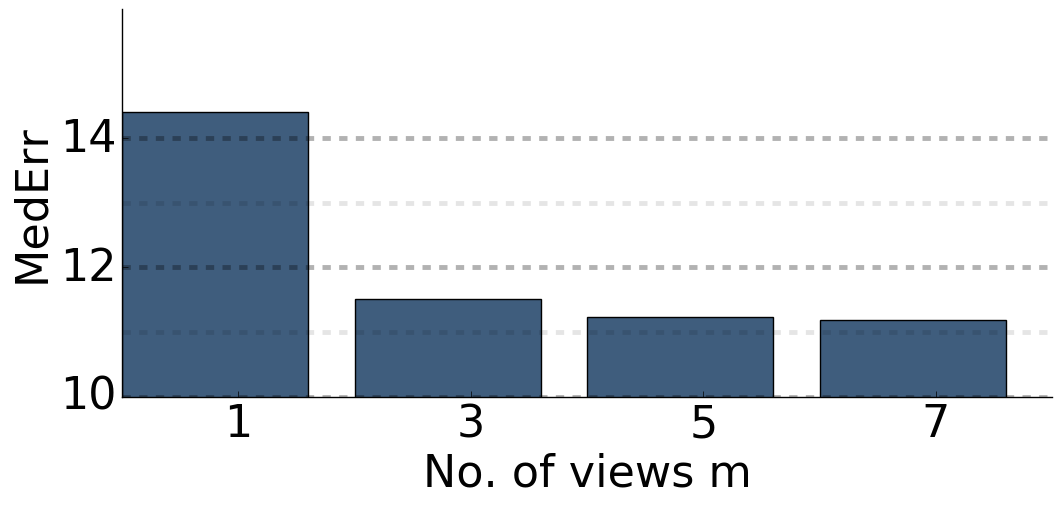}
}{%

  \caption{$Acc_{\pi/6}$ vs number for views '$m$' used for the multi-view information assimilation in our method.}%
  \label{fig:viewpoint}
}
\capbtabbox{%
  \begin{tabular}{l|cc}
 $\mathbf{\quad Ours_N \quad}$ &$\quad MedErr \quad$ & $\quad Acc_{\pi/6} \quad$\\ \hline
   w/o $L(I_2)$ & 11.51 & 0.74\\
  with $L(I_2)$ & \textbf{9.52} &\textbf{0.80}\\ \hline
  \end{tabular}
}{%
  \caption{Ablation on our model for validating the utility of $L(I_2)$ in improving pose estimation.}%
  \label{tab:ablation}
}
\end{floatrow}
\end{figure}

\subsection{Ablative Analysis}
\label{experiments:1}
In this section, we focus on evaluating the utility of various components of our method for object viewpoint estimation. Our ablative analysis focuses on the Chair category. The Chair category, having  high intra-class variation, is considered one of the most challenging classes and provides minimally biased dataset for evaluating ablations of our architecture. For all the ablations, the network is trained on the train-subset of ObjectNet3D and Pascal-3D+ dataset. We report our ablation statistics on the easy-test-subset of Pascal3D+ for chair category, as introduced by \cite{tulsiani2015viewpoints}.

First, we show the utility of the Multi-view information assimilation by performing ablations on the number of views `$m$'. In Figure \ref{fig:viewpoint}, we evaluate the $MedErr$ for our method with `$m$' varying from 1 to 7. Note that we do not utilize the local descriptors $L(I_2)$ in this setup and the pose estimator uses only the multi-view keypoint correspondence maps $mvK(I_2)$ as input. As the figure shows, additional information from multiple views is crucial. For having an computationally efficient yet effective system, we use $m = 3$ for all the following experiments. Next, it is essential to ascertain the utility of local descriptors $L(I_2)$ in improving our performance. In Table \ref{tab:ablation}, we can clearly observe increment in performance due to usage of $L(I_2)$ along with $mvK(I_2)$. Hence, in our final pipeline, the pose estimator network is designed to include the $L(I_2)$ as an additional input.

\subsection{Object Viewpoint estimation}

In this section, we evaluate our method against other \textit{state-of-the-art} approaches for the task of viewpoint estimation. Similar to other keypoint-based pose estimation works, such as 3D-INN~\cite{wu2016single}, we conduct our experiments on all object classes where 2D-keypoint information is available.

\noindent
\textbf{Pascal3D+}: Table \ref{tab:acc_pascal} compares our approach to other \textit{state-of-the-art} methods, namely Grabner~\etal~\cite{Grabner18} and RenderForCNN~\cite{su2015render}. The table shows, our best performing method $\mathbf{Ours_{D}}$ clearly outperform other established approaches on pose estimation task.

\begin{table*}[!t]
\centering
\begin{tabular}{l|cccccc}

\multirow{2}{*}{Category} \qquad &
 \multicolumn{2}{c}{Su~\etal~\cite{su2015render}} \qquad &\multicolumn{2}{c}{Grabner~\etal~\cite{Grabner18}} \qquad & \multicolumn{2}{c}{$\mathbf{Ours_D}$}   \\
& $Acc_{\pi/6}$ & \textit{MedErr} & $Acc_{\pi/6}$ & \textit{MedErr} & $Acc_{\pi/6}$ & \textit{MedErr}  \\ 
\hline
\hline
Chair &   0.86 & 9.7  & 0.80  & 13.7  & 0.83 & \textbf{8.84}  \\
Sofa  &  0.90  & \textbf{9.5}  & 0.87  & 13.5  & \textbf{0.90}  & 10.74   \\
Table & 0.73  & 10.8  & 0.71  & 11.8  & \textbf{0.87}  & \textbf{6.00}  \\ 
\hline
Average &  0.83  & 10.0  & 0.79  & 13.0  & \textbf{0.87}  & \textbf{8.53}  \\ 
\hline
\end{tabular}
\caption{Performance for object viewpoint estimation on PASCAL 3D+~\cite{xiang2014beyond} using ground truth bounding boxes. Note that $MedErr$ is measured in degree. }
\label{tab:acc_pascal}
\end{table*}

\noindent
\textbf{ObjectNet3D}: As none of the existing works have shown results on ObjectNet3D dataset, we trained RenderForCNN using the synthetic data and code provided by the authors Su~\etal~\cite{su2015render} for ObjectNet3D. Table~\ref{tab:acc_objnet} compares our method against RenderForCNN on various metrics for viewpoint estimation. RenderForCNN, which is trained using 500,000 more samples of synthetic images, still shows poor performance than the proposed method $\mathbf{Ours_N}$. 

\begin{table}[!b] 
\centering
\begin{tabular}{ccc|ccccc}
\setlength{\tabcolsep}{120pt}
Method& $\quad$ &Metric&$\quad$Chair$\quad$  & $\quad$Sofa$\quad$ & $\quad$Table$\quad$  & $\quad$Bed$\quad$ & $\quad$Avg.$\quad$ \\ 
\hline
\multicolumn{8}{c}{Object Viewpoint Estimation}\\
\hline
\hline
 \multirow{2}{*}{$MedErr$}&& Su~\etal~\cite{su2015render} &   9.70&  8.45 & 4.50 &  7.21 &  7.46 \\
&& $\mathbf{Ours_N}$&  \textbf{7.94} &  \textbf{3.55} & \textbf{3.33} &  \textbf{7.10} &  \textbf{5.48} \\ 
\hline
 \multirow{2}{*}{$Acc_{\pi/6}$}&& Su~\etal~\cite{su2015render} &  0.75 & 0.90 & 0.77 & 0.77 & 0.80 \\
&& $\mathbf{Ours_N}$& \textbf{0.81} &\textbf{0.92} &\textbf{0.90 }& \textbf{0.82} & \textbf{0.86} \\
\hline
 \multirow{2}{*}{$Acc_{\pi/8}$}&& Su~\etal~\cite{su2015render}& 0.71 & 0.89 &0.72 & 0.75 & 0.76 \\
&& $\mathbf{Ours_N}$& \textbf{0.78} & \textbf{0.90} & \textbf{0.88} & \textbf{0.79} & \textbf{0.83} \\
\hline
 \multirow{2}{*}{$Acc_{\pi/12}$}&& Su~\etal~\cite{su2015render}& 0.64 & 0.80 &0.68  &  0.72 & 0.71\\
&& $\mathbf{Ours_N}$ & \textbf{0.72} & \textbf{0.86} & \textbf{0.84}  &  \textbf{0.74} & \textbf{0.79} \\
\hline
\multicolumn{8}{c}{Joint Object Detection and Pose Estimation}\\
\hline
\hline
 \multirow{2}{*}{$AVP$-4}&& Su~\etal~\cite{su2015render}& \textbf{23.9} & 69.8 &53.5  &  65.1 & 53.1\\
&& $\mathbf{Ours_N}$& 22.1 & \textbf{71.9} &\textbf{65.7}  &  \textbf{71.6} &\textbf{ 57.8}\\
\hline
\end{tabular}
\caption{Evaluation on viewpoint estimation based tasks on the ObjectNet3D dataset. Note that $\mathbf{Ours_N}$ is trained with no synthetic data, where as Su~\etal is trained with 500,000 synthetic images (for all 4 classes).}
\label{tab:acc_objnet}
\end{table}

\subsection{Joint Object Detection and Viewpoint Estimation}
\noindent 
Now, for this task, our pipeline is used along with object detection proposal from R-CNN~\cite{girshick2014rich} using MCG~\cite{arbelaez2014multiscale} object proposals to estimate viewpoint of objects in each detected bounding box, as also followed by V\&K~\cite{tulsiani2015viewpoints}. Note that the performance of all models in this task is affected by the performance of the underlying Object Detection module, which varies significantly among classes.

\noindent
\textbf{Pascal3D+}: In Table \ref{tab:avp_pascal}, we compare our approach against other \textit{state-of-the-art} keypoint-based methods, namely, 3D-INN~\cite{wu2016single} and V\&K~\cite{tulsiani2015viewpoints}. 
The metric comparison shows superiority of our method, which in turn highlights ours’ ability to predict pose even with noisy  object localization. 

\begin{table}[!t] 
\centering
\begin{tabular}{l|cccc}
\setlength{\tabcolsep}{120pt}
$\quad$AVP$-$4$\quad$ & $\quad$Chair$\quad$  & $\quad$Sofa$\quad$ & $\quad$Table$\quad$  &  $\quad$Avg.$\quad$ \\ 
\hline
\hline
V\&K~\cite{tulsiani2015viewpoints}  &   25.1 &  43.8 &24.3 &  31.1 \\
\hline
3D-INN~\cite{wu2016single}  &   23.1 &  \textbf{45.8 }& - &   - \\
\hline
$\mathbf{Ours_D}$ & \textbf{26.0} & 41.9 &\textbf{26.5} & \textbf{31.5} \\
\hline
\hline
\end{tabular}
\caption{Comparison of $\mathbf{Ours_D}$ with other keypoint-based pose estimation approaches for the task of joint object detection and viewpoint estimation on Pascal3D+ dataset.}
\label{tab:avp_pascal}
\end{table}

\noindent
\textbf{ObjectNet3D}: Here, we trained RenderForCNN using the synthetic data and code provided by the authors Su~\etal~\cite{su2015render}. Table~\ref{tab:acc_objnet} compares our method against RenderForCNN on the $AVP$-$n$ metric.


Table~\ref{tab:acc_objnet} clearly demonstrates sub-optimal performance of RenderForCNN on ObjectNet3D. This is due to the fact that, the synthetic data provided by the authors Su~\etal~\cite{su2015render} is overfitted to the distribution of Pascal3D+ dataset. This leads to a lack of generalizability in RenderForCNN, where a mismatch in the synthetic and real data distribution can significantly lower its performance. Moreover, Table~\ref{tab:acc_objnet} not only presents our superior performance, but also highlights the poor generalizability of RenderForCNN.

\subsection{Analysis}
Here, we present analysis of results on additional experiments to highlight the chief benefits of the proposed approach.

\begin{table*}[!b]
\centering
\begin{tabular}{l|cccccc}

\multirow{2}{*}{Category} \qquad &
 \multicolumn{2}{c}{Su~\etal~\cite{su2015render}} \qquad & \multicolumn{2}{c}{Grabner~\etal~\cite{Grabner18}} \qquad & \multicolumn{2}{c}{ $\mathbf{Ours_N}$   }\\
& $Acc_{\pi/6}$ & \textit{MedErr} & $Acc_{\pi/6}$ & \textit{MedErr} & $Acc_{\pi/6}$ & \textit{MedErr}  \\ 
\hline\hline
Chair & 0.70 & 11.30  & 0.80  & 13.70  & \textbf{0.80}  & \textbf{9.52} \\
Sofa  & 0.65  & 14.45 &  \textbf{0.87}  & 13.50  & 0.80 &\textbf{ 9.96}   \\
Table & 0.70  & \textbf{5.80} & 0.71 & 11.80  & \textbf{0.83}  & 6.00 \\ 
\hline
Average &  0.68  & 10.51  & 0.79  & 13.0  & \textbf{0.81}  &\textbf{8.49 } \\ 
\hline
\end{tabular}
\caption{Performance for object viewpoint estimation on PASCAL 3D+~\cite{xiang2014beyond} using ground truth bounding boxes.
}
\label{tab:oursn}
\end{table*}
\noindent
\textbf{Effective Data Utilization}: To highlight the effective utilization of data in our method, we compare $\mathbf{Ours_{N}}$ against other methods trained without utilizing any synthetic data. For this experiment, we trained RenderForCNN without utilizing synthetic data and compare it to $\mathbf{Ours_N}$ in Table~\ref{tab:oursn}. The Table not only provides evidence for high data dependency of RenderForCNN, it also highlights our superior performance against Grabner~\etal~\cite{Grabner18} even in limited data scenario.

\begin{figure}[!t]
\begin{floatrow}\CenterFloatBoxes
\ffigbox[\FBwidth]{%
  
	\includegraphics[width=0.95\linewidth]{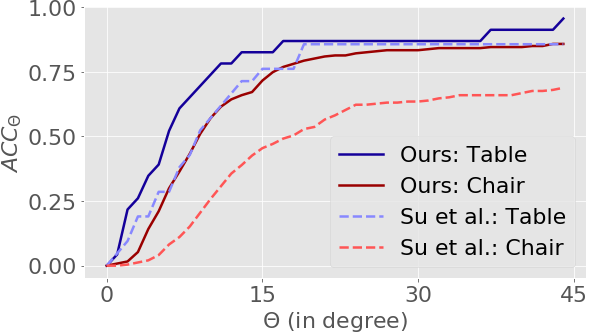}
}{%

  \caption{$Acc_{\theta}$ vs $\theta$ in Pascal3D+.} %
  \label{fig:fig_roc_pascal}
}
\ffigbox[\FBwidth]{%
  
	\includegraphics[width=0.95\linewidth]{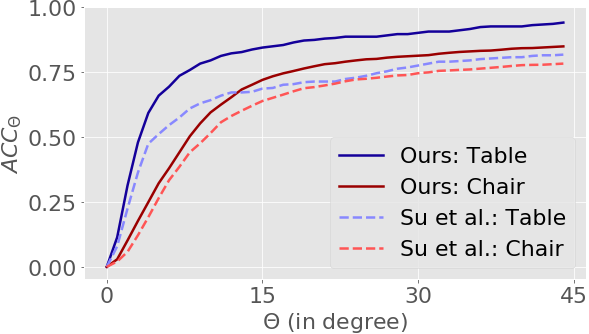}
}{%

  \caption{$Acc_{\theta}$ vs $\theta$ in ObjectNet3D.} %
  \label{fig:fig_roc_objnet}
}
\end{floatrow}
\end{figure}

\noindent
\textbf{Higher precision of our Approach}: Table~\ref{tab:higher} compares $\mathbf{Ours_N}$ to RenderForCNN~\cite{su2015render} on stricter metrics, namely $Acc_{\pi/8}$ and $Acc_{\pi/12}$. Further, we show a plot of $Acc_\theta$ vs $\theta$ in Figure  \ref{fig:fig_roc_pascal}, and \ref{fig:fig_roc_objnet} for multiple classes in both Pascal3D+ and ObjectNet3D dataset. 
Compared to the previous state-of-the-art model, we are able to substantially improve the performance with harsher $\theta$ bounds, indicating that our model is more precise on estimating the pose of objects on both 'Chair' and 'Table' category. This firmly establishing the superiority of our approach for the task of fine-grained viewpoint estimation.

\begin{table}[t]
\begin{tabular}{cc|cccc}
$\quad$ Metric $\quad$ & $\quad$ Method $\quad$ & $\quad$ Chair $\quad$ & $\quad$ Sofa $\quad$ & $\quad$ Table $\quad$ & $\quad$ Avg. $\quad$ \\ 
\hline
\hline
\multirow{3}{*}{$Acc_{\pi/8}$ } &Su~\etal~\cite{su2015render}&  0.59 & {\color{red}0.79}  & 0.68 &  0.68 \\
&$\mathbf{Ours_N} $   &  {\color{red}0.78} & 0.77 & {\color{red}0.83} &  {\color{red}0.79} \\
 &$\mathbf{Ours_D} $ & \textbf{0.81} & \textbf{0.85} & \textbf{0.86} &  \textbf{0.84} \\

\hline
\multirow{3}{*}{$Acc_{\pi/12}$ } &Su~\etal~\cite{su2015render}& 0.42&  {\color{red}0.69} &   0.60 &  0.57 \\
 &$\mathbf{Ours_N}$ & {\color{red}0.69}& 0.67 &   {\color{red}0.83} &  {\color{red}0.73} \\
&$\mathbf{Ours_D}$ &\textbf{ 0.72}&\textbf{ 0.75} & \textbf{0.83} &  \textbf{0.76} \\

\hline
\end{tabular}
  \caption{Comparison of our approach to existing \textit{state-of-the-art} methods for stricter metrics (On Pascal3D). For evaluating RenderForCNN on pascal3D+, the model provided by the authors Su~\etal has been used. The best value has been highlighted in \textbf{bold}, and the second best has been colored {\color{red}red}.}
  \label{tab:higher}
\end{table}

\section{Conclusions}

In this paper, we present a novel approach for object viewpoint estimation , which combines keypoint correspondence maps from multiple views, to achieve state-of-the-art results on standard pose estimation datasets. Being data-efficient, our method is suitable for large-scale or novel-object based real world applications. In future work, we would like to make the method weakly-supervised as obtaining keypoint annotations for novel object categories is non-trivial. Finally, the pose-invariant local descriptors show a promise of usability in other tasks, which will also be explored in the future.  


\bibliographystyle{splncs}
\bibliography{egbib}

\begin{thebibliography}{10}

\bibitem{mvRotationNet}
{Kanezaki}, A., {Matsushita}, Y., {Nishida}, Y.:
\newblock {RotationNet: Joint Object Categorization and Pose Estimation Using
  Multiviews from Unsupervised Viewpoints}.
\newblock ArXiv e-prints (March 2016)

\bibitem{mvtriplet}
{He}, X., {Zhou}, Y., {Zhou}, Z., {Bai}, S., {Bai}, X.:
\newblock {Triplet-Center Loss for Multi-View 3D Object Retrieval}.
\newblock ArXiv e-prints (March 2018)

\bibitem{mvhumanpose}
{Rhodin}, H., {Sp{\"o}rri}, J., {Katircioglu}, I., {Constantin}, V., {Meyer},
  F., {M{\"u}ller}, E., {Salzmann}, M., {Fua}, P.:
\newblock {Learning Monocular 3D Human Pose Estimation from Multi-view Images}.
\newblock ArXiv e-prints (March 2018)

\bibitem{xiang2014beyond}
Xiang, Y., Mottaghi, R., Savarese, S.:
\newblock Beyond pascal: A benchmark for 3d object detection in the wild.
\newblock In: WACV. (2014)

\bibitem{xiang2016objectnet3d}
Xiang, Y., Kim, W., Chen, W., Ji, J., Choy, C., Su, H., Mottaghi, R., Guibas,
  L., Savarese, S.:
\newblock Objectnet3d: A large scale database for 3d object recognition.
\newblock In: ECCV. (2016)

\bibitem{tulsiani2015viewpoints}
Tulsiani, S., Malik, J.:
\newblock Viewpoints and keypoints.
\newblock In: CVPR. (2015)

\bibitem{Grabner18}
Grabner, A., Roth, P.M., Lepetit, V.:
\newblock {3D Pose Estimation and 3D Model Retrieval for Objects in the Wild}.
\newblock In: {Proceedings of the IEEE Conference on Computer Vision and
  Pattern Recognition}. (2018)

\bibitem{su2015render}
Su, H., Qi, C.R., Li, Y., Guibas, L.J.:
\newblock Render for cnn: Viewpoint estimation in images using cnns trained
  with rendered 3d model views.
\newblock In: CVPR. (2015)

\bibitem{wu2016single}
Wu, J., Xue, T., Lim, J.J., Tian, Y., Tenenbaum, J.B., Torralba, A., Freeman,
  W.T.:
\newblock Single image 3d interpreter network.
\newblock In: ECCV. (2016)

\bibitem{aubry2014seeing}
Aubry, M., Maturana, D., Efros, A.A., Russell, B.C., Sivic, J.:
\newblock Seeing 3d chairs: exemplar part-based 2d-3d alignment using a large
  dataset of cad models.
\newblock In: CVPR. (2014)

\bibitem{liu2016sift}
Liu, C., Yuen, J., Torralba, A.:
\newblock Sift flow: Dense correspondence across scenes and its applications.
\newblock In: Dense Image Correspondences for Computer Vision.
\newblock Springer (2016)  15--49

\bibitem{taniai2016joint}
Taniai, T., Sinha, S.N., Sato, Y.:
\newblock Joint recovery of dense correspondence and cosegmentation in two
  images.
\newblock In: CVPR. (2016)

\bibitem{berg2005shape}
Berg, A.C., Berg, T.L., Malik, J.:
\newblock Shape matching and object recognition using low distortion
  correspondences.
\newblock In: CVPR. (2005)

\bibitem{schmidt2017self}
Schmidt, T., Newcombe, R., Fox, D.:
\newblock Self-supervised visual descriptor learning for dense correspondence.
\newblock IEEE Robotics and Automation Letters (2017)

\bibitem{han2017scnet}
Han, K., Rezende, R.S., Ham, B., Wong, K.Y.K., Cho, M., Schmid, C., Ponce, J.:
\newblock Scnet: Learning semantic correspondence.
\newblock In: ICCV. (2017)

\bibitem{yu2018hierarchical}
Yu, W., Sun, X., Yang, K., Rui, Y., Yao, H.:
\newblock Hierarchical semantic image matching using cnn feature pyramid.
\newblock Computer Vision and Image Understanding (2018)

\bibitem{choy2016universal}
Choy, C.B., Gwak, J., Savarese, S., Chandraker, M.:
\newblock Universal correspondence network.
\newblock In: NIPS. (2016)

\bibitem{Huang:2017:LMVCNN}
Huang, H., Kalogerakis, E., Chaudhuri, S., Ceylan, D., Kim, V.G., Yumer, E.:
\newblock Learning local shape descriptors from part correspondences with
  multiview convolutional networks.
\newblock ACM Transactions on Graphics \textbf{37}(1) (2017)

\bibitem{mv_1}
Borotschnig, H., Paletta, L., Prantl, M., Pinz, A.:
\newblock Appearance-based active object recognition.
\newblock Image and Vision Computing \textbf{18}(9) (2000)  715 -- 727

\bibitem{mv_2}
Paletta, L., Pinz, A.:
\newblock Active object recognition by view integration and reinforcement
  learning.
\newblock Robotics and Autonomous Systems \textbf{31}(1) (2000)  71 -- 86

\bibitem{MVCNN}
Su, H., Maji, S., Kalogerakis, E., Learned-Miller, E.:
\newblock Multi-view convolutional neural networks for 3d shape recognition.
\newblock In: Proceedings of the 2015 IEEE International Conference on Computer
  Vision (ICCV). ICCV '15, Washington, DC, USA, IEEE Computer Society (2015)
  945--953

\bibitem{mvOverview}
{Qi}, C.R., {Su}, H., {Niessner}, M., {Dai}, A., {Yan}, M., {Guibas}, L.J.:
\newblock {Volumetric and Multi-View CNNs for Object Classification on 3D
  Data}.
\newblock ArXiv e-prints (April 2016)

\bibitem{mvcTulsiani18}
Tulsiani, S., Efros, A.A., Malik, J.:
\newblock Multi-view consistency as supervisory signal for learning shape and
  pose prediction.
\newblock In: Computer Vision and Pattern Regognition (CVPR). (2018)

\bibitem{poirson2016fast}
Poirson, P., Ammirato, P., Fu, C.Y., Liu, W., Kosecka, J., Berg, A.C.:
\newblock Fast single shot detection and pose estimation.
\newblock In: 3DV. (2016)

\bibitem{mahendran20173d}
Mahendran, S., Ali, H., Vidal, R.:
\newblock 3d pose regression using convolutional neural networks.
\newblock In: ICCV. (2017)

\bibitem{kundu2018adadepth}
Kundu, J.N., Uppala, P.K., Pahuja, A., Babu, R.V.:
\newblock Adadepth: Unsupervised content congruent adaptation for depth
  estimation.
\newblock arXiv preprint arXiv:1803.01599 (2018)

\bibitem{szegedy2015going}
Szegedy, C., Liu, W., Jia, Y., Sermanet, P., Reed, S., Anguelov, D., Erhan, D.,
  Vanhoucke, V., Rabinovich, A.,  et~al.:
\newblock Going deeper with convolutions, CVPR (2015)

\bibitem{lowe2004distinctive}
Lowe, D.G.:
\newblock Distinctive image features from scale-invariant keypoints.
\newblock International journal of computer vision \textbf{60}(2) (2004)
  91--110

\bibitem{everingham2015pascal}
Everingham, M., Eslami, S.A., Van~Gool, L., Williams, C.K., Winn, J.,
  Zisserman, A.:
\newblock The pascal visual object classes challenge: A retrospective.
\newblock International journal of computer vision \textbf{111}(1) (2015)
  98--136

\bibitem{russakovsky2015imagenet}
Russakovsky, O., Deng, J., Su, H., Krause, J., Satheesh, S., Ma, S., Huang, Z.,
  Karpathy, A., Khosla, A., Bernstein, M.,  et~al.:
\newblock Imagenet large scale visual recognition challenge.
\newblock International Journal of Computer Vision \textbf{115}(3) (2015)
  211--252

\bibitem{kingma2014adam}
Kingma, D.P., Ba, J.:
\newblock Adam: A method for stochastic optimization.
\newblock arXiv preprint arXiv:1412.6980 (2014)

\bibitem{girshick2014rich}
Girshick, R., Donahue, J., Darrell, T., Malik, J.:
\newblock Rich feature hierarchies for accurate object detection and semantic
  segmentation.
\newblock In: Proceedings of the IEEE conference on computer vision and pattern
  recognition. (2014)  580--587

\bibitem{arbelaez2014multiscale}
Arbel{\'a}ez, P., Pont-Tuset, J., Barron, J.T., Marques, F., Malik, J.:
\newblock Multiscale combinatorial grouping.
\newblock In: CVPR. (2014)

\end{thebibliography}
\end{document}